\documentclass[conference, letterpaper]{IEEEtran}
\usepackage{cite}

\ifCLASSINFOpdf
\else
\fi
\usepackage{array}
\usepackage{url}

\usepackage{numprint} 
\usepackage{listings} 
\usepackage{color} 
\usepackage[dvipsnames]{xcolor}

\usepackage{colortbl} 
\definecolor{Gray}{gray}{0.9}
\newcolumntype{g}{>{\columncolor{Gray}}c}
\usepackage{multirow} 
\usepackage{booktabs} 

\usepackage{hyperref} 

\usepackage{subcaption}

\ifCLASSINFOpdf
   \usepackage[pdftex]{graphicx}
\else
\fi

\usepackage[cmex10]{amsmath}
\usepackage{color}
\usepackage{fancyhdr}

\renewcommand{\thispagestyle}[2]{}

\fancypagestyle{plain}{
        \fancyhead{}
        \fancyhead[C]{first page center header}
        \fancyfoot{}
        \fancyfoot[C]{first page center footer}
}
\begin{document}

%
\title{Applying advanced machine learning models to classify electro-physiological activity of human brain for use in biometric identification}

\author{\IEEEauthorblockN{Iaroslav Omelianenko}
\IEEEauthorblockA{NewGround LLC\\
Kiev, Ukraine\\
Email: yaric@newground.com.ua}}


%


\maketitle

\begin{abstract}
In this article we present the results of our research related to the study of correlations between specific visual stimulation and the elicited brain's electro-physiological response collected by EEG sensors from a group of participants. We will look at how the various characteristics of visual stimulation affect the measured electro-physiological response of the brain and describe the optimal parameters found that elicit a steady-state visually evoked potential (SSVEP) in certain parts of the cerebral cortex where it can be reliably perceived by the electrode of the EEG device. After that, we continue with a description of the advanced machine learning pipeline model that can perform confident classification of the collected EEG data in order to (a) reliably distinguish signal from noise (about 85\% validation score) and (b) reliably distinguish between EEG records collected from different human participants (about 80\% validation score). Finally, we demonstrate that the proposed method works reliably even with an inexpensive (less than \$100) consumer-grade EEG sensing device and with participants who do not have previous experience with EEG technology (EEG illiterate). All this in combination opens up broad prospects for the development of new types of consumer devices, [e.g.] based on virtual reality helmets or augmented reality glasses where EEG sensor can be easily integrated. 

The proposed method can be used to improve an online user experience by providing [e.g.] password-less user identification for VR / AR applications. It can also find a more advanced application in intensive care units where collected EEG data can be used to classify the level of conscious awareness of patients during anesthesia or to automatically detect hardware failures by classifying the input signal as noise.
\end{abstract}




%
\IEEEpeerreviewmaketitle


\section{Introduction}
A steady increase in the computing power of mobile devices allows them to be fully integrated with a variety of portable or wearable consumer devices such as virtual reality (VR) helmets, augmented reality (AR) glasses, monitors of the health and fitness, and so on. At the same time, they have enough computational resources to execute advanced machine learning models for processing of biometric data received from various sensors in the real time. One of these non-intrusive sensors that can be used to collect biometric information is the sensor to monitor the electro-physiological activity of the brain - electroencephalography (EEG). To collect EEG data it's enough to place electrodes along the scalp of the user. Moreover, the EEG sensors can be easily integrated into VR helmets or AR glasses, also providing easy means for specific visual stimulation. As has been demonstrated in recent studies \cite{Ding-Sperling-Srinivasan:2006, Regan:1997}, exposing a person to external stimuli, such as aural tone or a specific visualization, can cause a specific response in the electro-physiological activity of the brain. And this stationary cerebral cortex response, called steady-state visually evoked potentials (SSVEP) \cite{Regan:1997}, demonstrates a strong correlation with the parameters of visual stimuli, such as flicker frequency and visual appearance (configuration).

In \cite{Ding-Sperling-Srinivasan:2006}, it was found that a reliable SSVEP response can be detected in the Delta and Alpha (Low and High) frequency bands of the EEG signal collected at stimulation with a frequency of visual stimuli in the range: $4-10$Hz. That is, visual stimuli with a certain frequency and configuration can elicit a reliable SSVEP response in certain parts of the cerebral cortex, which leads to a resonance effect that can be easily detected by EEG sensors placed in specific positions on the scalp of the participant's head. (This can be partially explained if we consider neural networks of the cerebral cortex as a structural dynamic network of oscillators \cite{Tang-Basset:2017} which can be entrained by specific stimulus; where different regions of cerebral cortex having specific resonance frequencies amplifying its response on stimulation.) Later, the amplified response detected by EEG sensors can be processed using advanced machine learning classification methods to extract useful patterns.

In this work, we use a consumer-grade EEG sensing device with a single electrode located above the FP1 position, which collects the electrical activity of the frontal and pre-frontal cerebral cortex. Because of this limitation, we are most interested in finding specific visual stimuli capable of eliciting a reliable SVEP response in these areas. From the work of other researchers \cite{Ding-Sperling-Srinivasan:2006}, we know that the frontal cortex has a strong response in the High Alpha band, so we focused our efforts on creating a main visual stimulus with a frequency in the range of $8-10$Hz and auxiliary visual stimulus with the frequency of one harmonic down. In \cite{Ding-Sperling-Srinivasan:2006} it was found that with simultaneous use of several visual stimuli with frequencies related on the harmonic scale, the SSVEP response of certain parts of neural networks of the cerebral cortex can be amplified.

The purpose of this study is to create an advanced machine learning classification model suitable for processing of collected EEG records and capable of revealing the differences:
\begin{itemize}
\item[-] between \emph{signal} and \emph{noise},
\item[-] between EEG records collected from different human participants.
\end{itemize}

This paper is organized as follows: In Section~\ref{sec:experiment-setup-descr}, we describe the raw input signals received from the EEG device, visual stimulation details, and basic information about the involved participants. It is followed in Section~\ref{sec:data-corpus-description} by details about collected data records and how it was pre-processed before the analysis. In Section~\ref{sec:machine-learning-pipeline-description}, we provide description of machine learning classification model applied to the data corpus. Then, the Section~\ref{sec:finding-optimal-experiment-configuration} follows, describing the methodology used to find the optimal configuration of visual stimulation and to find optimal hyper-parameters of the machine learning classification model. After that in Section~\ref{sec:results} we consider achieved results (validation scores) for both signal / noise and participants identification tasks. Finally, in Section~\ref{sec:future-works} and ~\ref{sec:conclusion} we describe our plans for future experiments and review the results obtained during the research.


\section{Experiment Setup Description}\label{sec:experiment-setup-descr}


\subsection{EEG monitoring device and raw input signals}

In our experiments, we use NeuroSky MindWave device that monitors the brain's electro-physiological activity using one electrode at the \emph{FP1} position and other at the ear lobe. It has a sampling rate of $512$ Hz and decomposes the raw input signal on several frequency bands at the hardware level using a fast Fourier transform (FFT) \cite{Heideman-Johnson-Burrus:1984}. After FFT decomposition, the MindWave device broadcasts the decomposed EEG signal once per second (i.e., the refresh rate is $1$Hz). See Table~\ref{tbl:eeg_bands}

\begin{table}[ht]
\renewcommand{\arraystretch}{1.3}
\caption{The ranges of frequency bands for FFT decomposed EEG signal}
\label{tbl:eeg_bands}
\centering
\begin{tabular}{lc}
\toprule
Band name & Frequency range, Hz \\
\midrule
Delta, \(\Delta\) & 1-3 \\ \hline
Theta, \(\Theta\) & 4-7\\ \hline
Low Alpha, \(L\alpha\) & 8-9\\ \hline
High Alpha, \(H\alpha\) & 10-12\\ \hline
Low Beta, \(L\beta\) & 13-17\\ \hline
High Beta, \(H\beta\) & 18-30\\ \hline
Low Gamma, \(L\gamma\) & 31-40\\ \hline
High Gamma, \(H\gamma\) & 41-50\\
\bottomrule
\end{tabular}
\end{table}

Additionally, it provides two \emph{complex synthetic signals} calculated by combining data from several frequency bands, namely:

\begin{itemize}
\item[-] \emph{Attention} to indicate user’s attention level in range $[1, 100]$
\item[-] \emph{Meditation} to indicate user’s contemplation level in range $[1, 100]$
\end{itemize}

We use the mentioned \emph{complex synthetic signals} as control signals that trigger the recording / processing of EEG data when a certain threshold value is reached.


\subsection{Visual stimulation and data collection}\label{sec:visual-stimulation-and-data-collection}

Following research \cite{Ding-Sperling-Srinivasan:2006}, we created an advanced software library to produce a certain visual stimulation. The visualization library provides tools for configuring various visualization options and applying various visual modeling schemes: \emph{WISP} and \emph{WAVE}. This allows a series of experiments with different parameters to be performed to find the optimal parameters for eliciting proper SSVEP response.

\begin{figure}[t]
\centering
\includegraphics[width=3in]{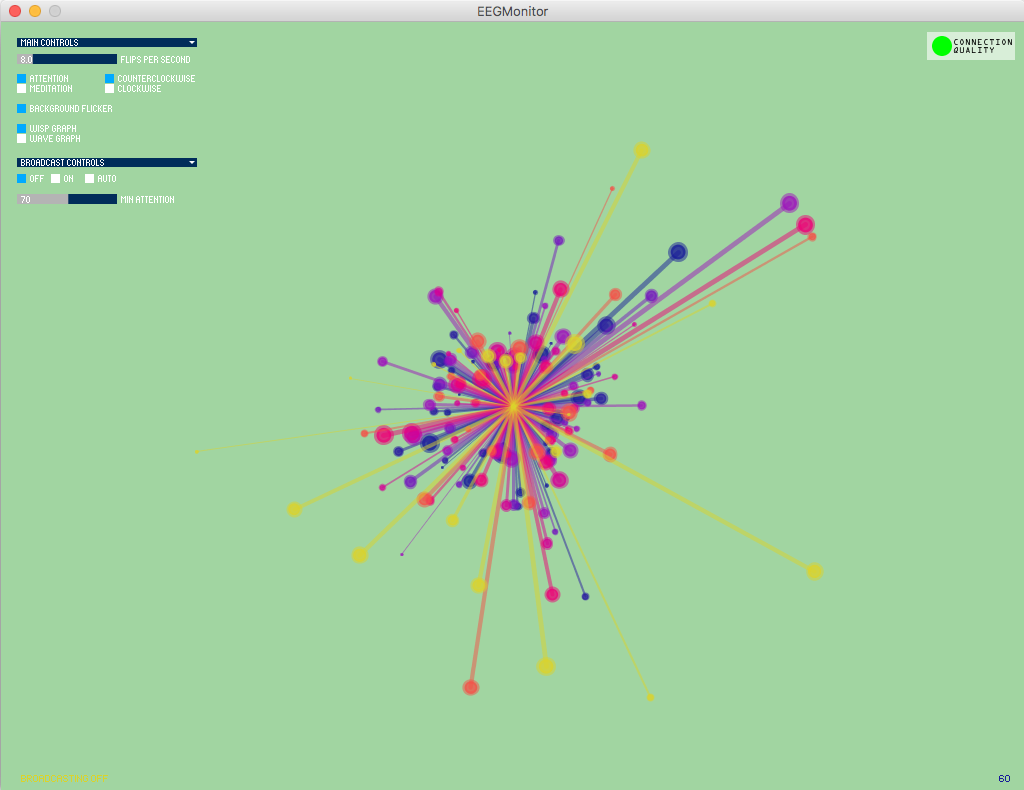}
\caption{The screen shoot of visual stimulator window in \emph{WISP} rendering mode.}
\label{fig:wisp_screen}
\end{figure}

The visual stimuli created by the visualization library have a strong correlation with the collected EEG data in real time, providing positive feedback to the user on the activity of the cerebral cortex. In the \emph{WISP} rendering mode, this correlation is represented as the relative position of points in polar coordinates and as the color of points and wisp lines (see Figure~\ref{fig:wisp_screen}). The position of the points relative to the center, determined by the strength of the EEG signal for a particular frequency band. Each frequency band has a certain color assigned to it. To elicit appropriate SSVEP response the visualization has two flickering areas: 
\begin{enumerate}
\item wisp of lines with dots - flickers with selected primary frequency
\item background - flickers with secondary frequency (one harmonic down)
\end{enumerate}

The background color saturation is determined by the current value of selected control signal - \emph {Attention} or \emph {Meditation}, where higher values result in a more saturated color. Thus, the saturation of the background color provides positive feedback for the participant about current value of the control signal.

The central idea of the described scheme of visual stimulation is to achieve aesthetically pleasing and attractive visualization, which can be used in consumer devices: helmets VR, AR glasses, etc. But at the same time, it should provide appropriate visual stimulation to elicit proper SSVEP response from cerebral cortex.

\begin{figure}[t]
\centering
\includegraphics[width=3in]{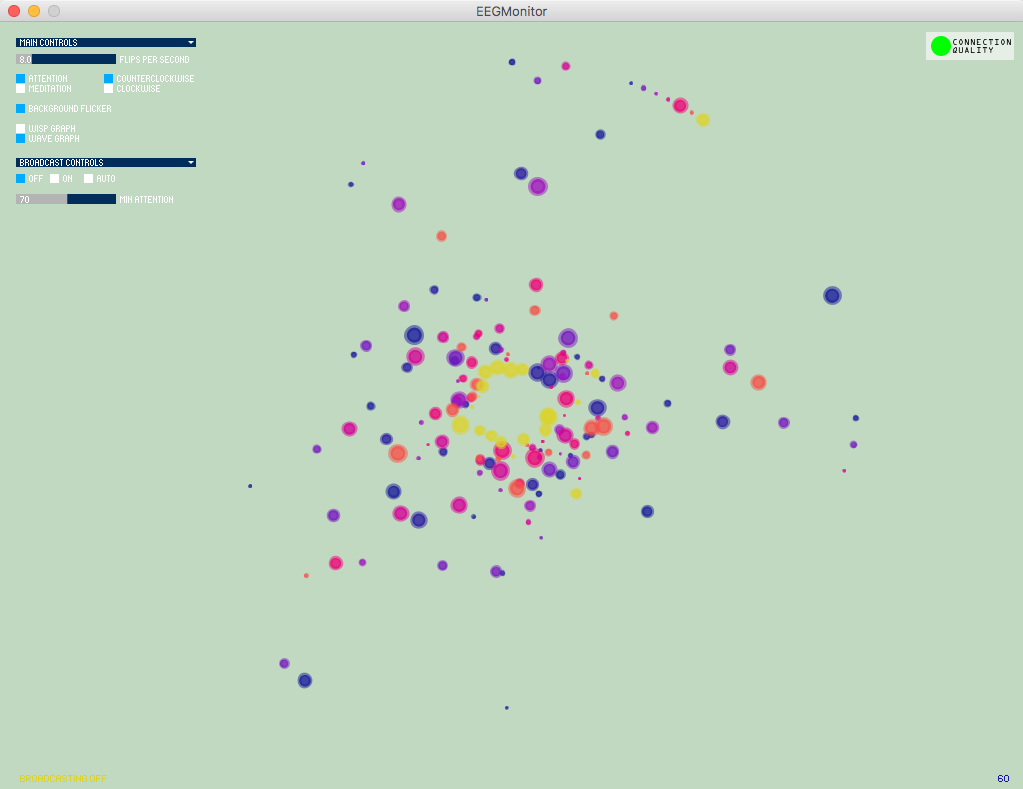}
\caption{The screen shoot of visual stimulator window in \emph{WAVE} rendering mode.}
\label{fig:wave_screen}
\end{figure}

Another studied mode of visual stimulation, we called \emph{WAVE}. Its main difference from \emph{WISP} is that it displays only dots, skipping lines that connect them to the center point (see Figure~\ref{fig:wave_screen}). We found that different visual modes have a variable effect on elicited SSVEP response in different frequency ranges of stimuli. In the \emph{WISP} rendering mode the most effective primary stimulation frequency is \(10\)Hz (High Alpha) and secondary - \(5\)Hz. And for the \emph{WAVE} this is: primary - \(8\)Hz (Low Alpha) and secondary - \(4\)Hz.

The graphic interface of the created visualization library also allows you to set a certain control signal threshold that starts / stops recording the received EEG signals. We learned that with higher threshold values, it is much more difficult for a participant to maintain that level of control signal (\emph{Attention} / \emph{Meditation}), but the collected data is of better quality. Thus, it is important to find the right balance between the quality of the collected data and ease of use for the participants. In our experiments, we applied four threshold values from $80$ down to $50$ in steps of ten. The optimal threshold value for the \emph{WISP} rendering mode is $70$ providing a good balance between the quality of the collected data and ease of use for the participant.

We measure quality of the collected data and the applicability of a specific configuration of visual stimuli, by checking Pearson product-moment correlation \cite{Gain:1951} between the processed data from the recorded sessions for each participant per each configuration. (For more information on calculating correlation coefficients, see Section: "~\nameref{sec:optimal-configuration-visual-stimulation}".) And for further analysis, the configuration of visual stimuli creating the most correlated results is considered. Our hypothesis for this is that the maximum correlation between recording sessions is the result of properly elicit SSVEP response from a specific parts of the cerebral cortex that has a steadily recognizable pattern among all recorded sessions.

Applying the above-mentioned method to estimate quality of the collected EEG data, following visualization parameters were chosen for all participants as optimal:
\begin{itemize}
\item[-] \emph{WISP} rendering mode
\item[-] $10$Hz primary and $5$Hz secondary flicker frequencies
\item[-] \emph{Attention} complex synthetic signal as a control signal
\item[-] $70$ as the control signal threshold value
\end{itemize}

In a series of experiments, we found that reliable data processing requires at least \emph{forty} collected data samples (EEG events) per session. The MindWave device broadcasts data event signals (with $8$ frequency bands values) once per second. This means that one recording session should take at least $40$ seconds, if participant can maintain a continuous high-level value of the control signal. But this can take much longer if the value of the control signal drops below the threshold and the recording pauses until participant can again reach the threshold value.


\subsection{Experiment Participants}

In our experiments we collected EEG data records from three participants:
\begin{enumerate}
\item male, $43$ years, $-1$ myopia, astigmatism
\item female $28$ years, normal sight
\item male $35$ years, $-7$ myopia
\end{enumerate}

For each participant, we collected from six to eighteen sessions of EEG records with various schemes of visual stimulation and at different times. It is very important that EEG sessions be recorded at different times in order to observe the electro-physiological response of the participants' brain in different conditions.


\section{Data Corpus Description}\label{sec:data-corpus-description}


The data collected from the EEG device during a particular recording session is stored as comma-separated records with $40$ rows, $11$ columns wide. Each line contains three meta-data variables (time stamp of record, frequency of primary and secondary stimuli) and eight values for each frequency band.

The frequency bands data collected from EEG device represent a power spectrum and its values vary exponentially in the range $[0, 32767]$. But performance of a majority of machine learning methods is best with small floating point values centered around zero. Thus, in order to improve the quality of the collected data and to balance the values of the data points for different frequency bands, we scaled it to fit in range $[0, 1]$ during the preprocessing. And the pre-processed data is later used as input for the machine learning classification model.


\section{Machine Learning Classification Pipeline Model Description}\label{sec:machine-learning-pipeline-description}


Essentially, the collected data corpus of EEG records is a time series. Therefore, it may seem natural to analyze it using machine learning methods, which are usually considered to have the best architecture for time series data analysis, e.g. Recurrent Neural Networks (RNN) and its modern reincarnation - Long Short Time Memory Neural Network (LSTM) \cite{Hochreiter-Sepp-Schmidhuber-Jürgen:1997}. But for RNN model to learn useful patterns from input data at least two conditions should be met: (a) the data samples should be collected at equal time intervals between events; (b) it is important to collect a significant number of data samples for each recording session. In our experiment, no above condition can be fulfilled. First of all, it is unpractical to collect a lot of data samples for consumer device setup, because people easily get bored during long recording sessions. Furthermore, we collect EEG data events only when the level of the control signal exceeds a certain threshold, which makes our time series rather inconsistent with different intervals between time stamps of collected events.

\begin{figure}[t]
\centering
\includegraphics[width=3in]{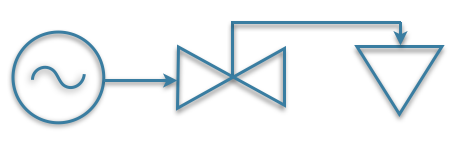}
\caption{The machine learning processing pipeline scheme from left to right: the EEG input data source, the auto-encoder, and the classifier.}
\label{fig:setup_ml_scheme}
\end{figure}

Taking into account the specific conditions of our experiment mentioned above, we decided to explore a novel approach to the analysis of time series, which allows us to extract useful patterns (essence) from the analyzed EEG signal records and classify it later. Our idea is to create advanced machine learning classification pipeline, where the input EEG data is processed in two stages (see Figure~\ref{fig:setup_ml_scheme}):
\begin{itemize}
\item[-] at the first stage, raw input data are analyzed using unsupervised machine learning method based on the auto-encoder neural network with single hidden layer \cite{Bengio:2009}
\item[-] at the second stage, the learned weights of the auto-encoder hidden layer (data encoding) are used as input for the classifier of choice to train the supervised classification model
\end{itemize}

In the following we describe considered methods of machine learning used at both stages, with more details.


\subsection{The First stage - Auto-Encoder Selection}

In our experimental setup - per each individual session - we have the data corpus consisting of eight frequency bands values collected for forty data events. That gives us the features space dimension width of $320$ for each EEG recording session. At the same time, we have very limited number of recording sessions ($6$ to $18$) for every participant, which is much less than the dimensionality of the features space. Such characteristics of the input data make it unacceptable for analysis by almost any machine learning classification model, since ideally the number of data samples must significantly exceed the number of features and not vice versa. This forced us to look for a specific method of machine learning that can extract its reliable representation (encoding) from the raw EEG input signal and accordingly reduce the dimensionality of the features space.

Our attention was captured by family of unsupervised machine learning methods called auto-encoders \cite{Bengio:2009}. An auto-encoder takes an input \(\mathbf{x} \in [0,1] ^{d}\) and first maps it (with an encoder) to a hidden representation \(\mathbf{h} \in [0,1] ^{d'}\) through a deterministic mapping, e.g.:
\begin{equation}
\label{eqn:auto_hidden_mapping}
\mathbf{h} = \sigma(\mathbf{W}\mathbf{x} + \mathbf{b})
\end{equation}where \(\sigma\) is a non-linearity such as the sigmoid.

The latent representation \(\mathbf{h}\), or \textbf{encoding} is then mapped back with a decoder into a reconstruction \(\mathbf{z}\) of the same shape as \(\mathbf{x}\). The mapping happens through a similar transformation, e.g.:
\begin{equation}
\label{eqn:auto_decoder_mapping}
\mathbf{z} = \sigma(\mathbf{W'}\mathbf{h} + \mathbf{b'})
\end{equation}\(\mathbf{z}\) should be seen as a prediction of \(\mathbf{x}\), given the code \(\mathbf{h}\). Optionally, the weight matrix \(\mathbf{W'}\) of the reverse mapping may be constrained to be the transpose of the forward mapping: \(\mathbf{W'} = \mathbf{W}^T\). This is referred to as tied weights. The parameters of this model (\(\mathbf{W}\), \(\mathbf{b}\), \(\mathbf{b'}\) and, if one doesn't use tied weights, also \(\mathbf{W'}\)) are optimized such that the average reconstruction error is minimized.

The reconstruction error can be measured in many ways, depending on the appropriate distributional assumptions on the input given the code. The traditional squared error \(L(\mathbf{x} \mathbf{z}) = || \mathbf{x} - \mathbf{z} ||^2\), can be used as well. If the input is interpreted as either bit vectors or vectors of bit probabilities, the cross-entropy of the reconstruction can be used:
\begin{equation}
\label{eqn:auto_crosss_entropy}
L_{H} (\mathbf{x}, \mathbf{z}) = - \sum^d_{k=1}[\mathbf{x}_k \log \mathbf{z}_k + (1 - \mathbf{x}_k)\log(1 - \mathbf{z}_k)]
\end{equation}

The \(\mathbf{h}\) can be viewed as lossy compression of \(\mathbf{x}\) and optimization makes it a good compression for training examples and hopefully for all other examples as well. But this can not be guaranteed for arbitrary inputs. The simple auto-encoder gives low reconstruction error on test samples from the same distribution as train samples, but usually gives higher error values for inputs randomly chosen from input space.

To prevent auto-encoder from learning identity function of input signal and to force it to learn something useful about input signal in its hidden units, several methodologies can be applied: (a) addition of sparsity to inputs, (b) addition of randomness to transformation (encoding) or (c) introducing specific explicit regularizer to the objective function.

In this research we study two types of auto-encoders with various learning enhancements applied:
\begin{itemize}
\item[-] Denoising auto-encoder (addition of sparsity to inputs)
\item[-] Contractive auto-encoder (the explicit regularizer to the objective function)
\end{itemize}

The denoising auto-encoder \cite{Vincent-Larochelle-Bengio-Manzagol:2008} is a stochastic version of simple auto-encoder where stochastic corruption process randomly sets many of inputs (maximum half of them) to zero. Then we train it to reconstruct correct input from a stochastically corrupted version of it. As result, the higher level representations of input data (encoding) are relatively stable and robust to the corruption of input data.

The contractive auto-encoder \cite{Rifai-Vincent-Muller-Glorot-Bengio:2011} adds explicit regularizer to the objective function that forces model to obtain a robust representation of the input space not sensitive to slight variations of input data. This regularizer corresponds to the Frobenius norm (L2) of the Jacobian matrix of the hidden representation with respect to the input. The final objective function has the following form:
\begin{equation}
\label{eqn:auto_contractive}
L = -\sum_{k=1}^d [\mathbf{x}_k \log \mathbf{z}_k + (1-\mathbf{x}_k) \log( 1-\mathbf{z}_k)] + \lambda \sum_{i=1}^d \sum_{j=1}^n \mathbf{J}_{ij}^2
\end{equation}where \(\mathbf{z} = \sigma(\mathbf{W}' \mathbf{h}  + \mathbf{b}')\) is the reconstruction of input vector, \(\mathbf{J}_i = \mathbf{h}_i (1 - \mathbf{h}_i)\cdot \mathbf{W}_i\) is Jacobian of \(\mathbf{h}\) with respect to \(\mathbf{x}\), and \(\mathbf{h} = \sigma(\mathbf{W}\mathbf{x} + \mathbf{b})\) is the projection of the input into the latent space \(\mathbf{h}\).

Both mentioned auto-encoder variations are robust to the slight corruption of input data. This is especially useful for our experiment, since it is well known that the electro-physiological activity of the brain is non-stationary and varies greatly, depending on the circumstances. I.e. the collected EEG data for each subsequent recording session under the same type of visual stimuli can give the same essence (compressed representation), but can introduce various non-stationary deviations from previous sessions. Using an auto-encoder to analyze the input data at the first stage, we hope to reduce non-stationary deviations and extract a useful compressed representation. Then, at the second stage, the extracted compressed representation with significantly reduced dimensionality of features space will be fed into the classifier of choice.

The above-mentioned auto-encoders was implemented in Python programming language using Theano framework \cite{Rami_Al-Rfou:2016}.


\subsection{The Second Stage - Classifiers Evaluation}\label{sec:sec-stage-classifiers-evaluation}

At the end of the first stage of the processing pipeline, the dimensionality of data corpus features space is considerably reduced. After that, at the second stage, the data is fed to a set of classifiers from simple to advanced in order to find the one that has the most powerful prediction model.

To automate the evaluation of classifiers against different sets of hyper-parameters, we use exhaustive grid search with cross-validation. This method exhaustively generates candidates from the grid of parameter values specified for each classifier, and then performs a $3$\emph{-fold} cross-validation (CV) of the predictive performance of each classifier with respect to the input data \cite{Kohavi:1995}. In the k-fold cross-validation method, the training set is split into smaller sets and the following procedure is performed for each of the folds:
\begin{enumerate}
\item a model is trained using \(k - 1\) of the folds as training data
\item the resulting model is validated on the remaining part of the data (i.e., it is used as a test set to compute a performance measure such as accuracy)
\end{enumerate}

The resulting classifier performance measure is done by averaging scores computed for each fold. Thus, we can avoid the effect of model overfitting, when it shows perfect score for train data but fails to predict anything useful for unseen data. Refer to Table~\ref{tbl:tested_classifiers} for a list of all tested classifiers and all tested values of corresponding hyper-parameters per each estimator.

Hereafter we provide short description of each tested classifier.

\begin{table}[ht]
\renewcommand{\arraystretch}{1.3}
\caption{The classifiers used in the study with range of hyper-parameter values applied for exhaustive grid search.}
\label{tbl:tested_classifiers}
\centering
\begin{tabular}{p{1in}|p{2in}}
\toprule
Classifier & Parameters space \\
\midrule
Random Forest & max\_depth: [3, 5, 8, None], max\_features: [1, 3, 4], min\_samples\_split: [2, 3, 4], min\_samples\_leaf: [1, 3, 10], bootstrap: [True, False], criterion: ["gini", “entropy”], n\_estimators: [10, 20, 50, 100, 200]\\ \hline
Ada Boost & learning\_rate: [0.01, 0.1, 1], n\_estimators: [10, 20, 50, 100]\\ \hline
Decision Tree & max\_depth: [3, 5, 8, None], min\_samples\_split: [2, 3, 4], min\_samples\_leaf: [1, 3, 10, 20, 30]\\ \hline
Gaussian Process & warm\_start: [True, False]\\ \hline
Multi Layer Perceptron & alpha: [0.0001, 0.001, 0.01], learning\_rate\_init: [0.001, 0.01, 0.1, 0.5], momentum: [0.9, 0.99, 0.999], solver: ["lbfgs", "sgd", “adam"], activation: ["logistic", "tanh", “relu"], hidden\_layer\_sizes: [4, 8, 10]\\ \hline
K-neighbors & n\_neighbors: [2, 3, 5], algorithm: ["ball\_tree", "kd\_tree", "brute"]\\ \hline
Gaussian Naïve Bayes & priors: [None]\\ \hline
Quadratic Discriminant Analysis & priors: [None], reg\_param: [0.0, 0.01, 0.1, 0.9]\\ \hline
Support Vector Machine with RBF kernel & C: [0.1, 0.5, 1.0], gamma: [0.1, 0.5, 1.0, 2.0, 3.0, 'auto']\\ \hline
Support Vector Machine with Linear kernel & C: [0.001, 0.005, 0.01, 0.025, 0.05, 0.1, 0.25, 0.5, 1.0]\\
\bottomrule
\end{tabular}
\end{table}

\textbf{Random Forest classifier} \cite{Breiman:2001} is a meta estimator that fits a number of decision tree classifiers on various sub-samples of the dataset and use averaging to improve the predictive accuracy and control overfitting.

\textbf{Ada Boost classifier} \cite{Zhu-Zou-Rosset-Hastie:2009} is a meta-estimator that begins by fitting a classifier on the original dataset and then fits additional copies of the classifier on the same dataset but where the weights of incorrectly classified instances are adjusted such that subsequent classifiers focus more on difficult cases.

\textbf{Decision Tree classifier} \cite{Breiman-Friedman-Olshen-Stone:1984} is a flow-chart-like structure, where each internal (non-leaf) node denotes a test on an attribute, each branch represents the outcome of a test, and each leaf (or terminal) node holds a class label. This structure allows to go from observations (branches) to the conclusions about the item's target value (leaves).

\textbf{Gaussian Process classifier} \cite{Rasmussen-Williams:2006} based on Laplace approximation, which is used for approximating the non-Gaussian posterior by a Gaussian.

\textbf{Multi-layer Perceptron (MLP) classifier} \cite{Hinton:1989} is an artificial neural network based classifier optimizing log-loss function using LBFGS or stochastic gradient descent.

\textbf{K-neighbors classifier} \cite{Altman:1992} use a type of instance-based learning or non-generalizing learning where classification is computed from a simple majority vote of the nearest neighbors of each point: a query point is assigned the data class which has the most representatives within the nearest neighbors of the point.

\textbf{Gaussian Naive Bayes} \cite{Chan-Golub-LeVeque:1979} is a model assuming that the continuous values associated with each class are distributed according to a Gaussian distribution.

\textbf{Quadratic Discriminant Analysis} \cite{Hastie-Tibshirani-Friedman:2008} is a classifier with a quadratic decision boundary, generated by fitting class conditional densities to the data and using Bayes’ rule.

\textbf{Support Vector Machine classifier} \cite{Cortes-Vapnik:1995} builds a representation of the examples as points in space, mapped so that the examples of the separate categories are divided by a clear gap that is as wide as possible. New examples are then mapped into that same space and predicted to belong to a category based on on which side of the gap they fall.

We implemented exhaustive grid search with $3$\emph{-fold} cross-validation for all above mentioned classifiers in Python programming language using Scikit-learn framework \cite{Pedregosa-et-al:2011}.


\section{Finding Optimal Experiment Configuration}\label{sec:finding-optimal-experiment-configuration}


The details of the experiment described above suggest that in order to obtain reliable classification, it is important to find optimal configuration parameters for both parts of the experiment: visual stimulation and machine learning classification pipeline.

Hereafter we consider in more details how optimal configuration was found.


\subsection{Finding Optimal Configuration for Visual Stimulation}\label{sec:optimal-configuration-visual-stimulation}

To achieve the objectives of the experiment, first, we need to find the optimal configuration of visual stimuli that elicit a reliable SSVEP response in the cerebral cortex of participants. The optimal configuration should give comparable results for different EEG sessions recorded on different times. As mentioned in Subsection:~"\nameref{sec:visual-stimulation-and-data-collection}" to do this we conducted series of experiments (recording sessions) and evaluate collected data by calculating Pearson product-moment correlation \cite{Gain:1951} between outputs from the first stage of machine learning pipeline (auto-encoder). Our hypothesis is that the visual stimulation scheme leading to outputs with the maximal correlation can be considered optimal.

\begin{figure}[t]
\centering
\includegraphics[width=3in,keepaspectratio=true]{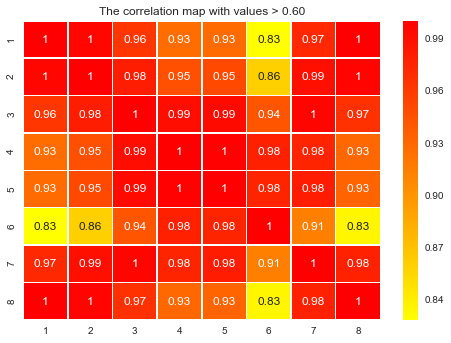}
\caption{The heat map with correlation coefficients between eight recording sessions for one of the experiment participants. It can be seen that majority of sessions highly correlate with each other and only session \#$6$ is a bit outlier. Such high correlation values is a good sign that applied configuration of the visual stimuli is optimal to elicit robust SSVEP response which can be compressed to extract useful pattern (encoding).}
\label{fig:corr-heat-map}
\end{figure}

The correlation coefficients can be calculated as following:
\begin{enumerate}
\item The collected EEG raw input data preprocessed by scaling it down to fit range $[0,1]$  (see Section:~"\nameref{sec:data-corpus-description}").
\item After that, the scaled data corpus is fed into auto-encoder of choice. The input data is filtered to include only \emph{Delta} and \emph{High Alpha} bands. These bands were chosen based on the results of \cite{Ding-Sperling-Srinivasan:2006}, which state that the elicited SSVEP response should have bands frequencies close to the frequencies of the visual stimuli ($10/5$Hz). It was found that for best results the auto-encoder should have only two units in the hidden layer.
\item The learned weight matrix of the auto-encoder's hidden layer (compressed input data representation) per each recording session then collected as representation data matrix with columns holding compressed data for each session.
\item Finally, the Pearson product-moment correlation coefficients calculated from representation matrix and  heat map plot with correlations are rendered (see Figure~\ref{fig:corr-heat-map}). 
\end{enumerate}


\subsection{Finding Optimal Configuration for Machine Learning Classification Pipeline Model}

The optimal configuration of classification pipeline was found in a series of experiments by applying various hyper-parameters to auto-encoders and classifiers. After that, the classification model with the maximum validation score was chosen as optimal. To automate the process of selecting the optimal model, a software framework was created that performs the following procedures:
\begin{enumerate}
\item starts preprocessor to process the raw input data corpus consisting of collected EEG recording sessions,
\item uses auto-encoder against preprocessed input data to extract useful patterns,
\item perform classifiers evaluation against extracted patterns as described in Section:~"\nameref{sec:sec-stage-classifiers-evaluation}".
\end{enumerate}

The experimental framework allows to separately define various configuration parameters for each processing stage. Thus, the configuration of each experiment can be encapsulated in a separate Python script, which makes it easy to reproduce a specific experiment in the future.

We tested a variety of hyper-parameters combinations per each auto-encoder with different sets of filtered input data samples (only values for specific frequency bands are included). As result, we found that learning rate, contraction / corruption level and the number of learning epochs can be fixed by the values presented in Table~\ref{tbl:hyper-parameters-auto-encoders}. Only the batch size, the number of auto-encoder's hidden units and frequency bands included into input data corpus, can be changed when searching for the best validation score.

\begin{table}[ht]
\renewcommand{\arraystretch}{1.5}
\caption{The optimal hyper-parameters for auto-encoders for signal / noise classification task.}
\label{tbl:hyper-parameters-auto-encoders}
\centering
\begin{tabular}{m{1in}m{0.6in}m{0.6in}m{0.6in}}
\toprule
Hyper-parameter & Tested values range & Contractive auto-encoder & Denoising auto-encoder \\
\midrule
Batch size & $[1,10]$ & 1 or 10 & 5\\ \hline
Learning rate & $[0.001,0.1]$ & 0.1 & 0.1\\ \hline
Number of units in hidden layer & $[2,10]$ & 5 & 5\\ \hline
Number of learning epochs & $[10^4,5\cdot10^4]$ & $5\cdot10^4$ & $5\cdot10^4$\\ \hline
Contraction level (contractive auto-encoder) & $[0.1,0.3]$ & 0.1 & -\\ \hline
Corruption level (denoising auto-encoder) & $[0.1,0.3]$ & - & 0.1\\ \hline
Included frequency bands & $\Delta$, $\Theta$, L$\alpha$, H$\alpha$, L$\beta$, H$\beta$, L$\gamma$, H$\gamma$ & $\Delta$, $\Theta$, L$\alpha$, H$\alpha$, L$\beta$, H$\beta$ & $\Delta$, $\Theta$, L$\alpha$, H$\alpha$, L$\beta$, H$\beta$\\
\bottomrule
\end{tabular}
\end{table}


\section{Results}\label{sec:results}


In this paper, we considered the possibility of creating reliable machine learning classification pipeline model able to:
\begin{enumerate}
\item confidently distinguish between signal and noise in collected EEG records
\item confidently classify EEG records of different participants, that is, provide means to determine which EEG record belongs to which participant.
\end{enumerate}

Hereafter we consider achieved results in more details.


\subsection{Signal / Noise Classification}

For this task our main goal is to build classifier capable of distinguishing between \emph{signal} and \emph{noise} records collected under different conditions and at different times. The \emph{signal} records include SSVEP affected records plus EEG records obtained from participants performing other cognitive tasks (sky watching, reading, etc). The basic requirement for the proper \emph{signal} record - is to be collected when the value of the control signal (Attention or Meditation) is greater than or equal to the specified threshold value. The \emph{noise} records are collected when the electrodes of the EEG device are improperly installed on the participants (for example, the base electrode is not connected) or when the device is turned on but not connected to the participant at all. At total, we studied data samples collected over $26$ \emph{signal} and $26$ \emph{noise} sessions.

\begin{table}[ht]
\renewcommand{\arraystretch}{1.5}
\caption{The comparison of the best validation scores for signal / noise classification per particular set of included frequency bands and specific values of contractive auto-encoder hyper-parameters. The score value is given as the average of $3$\emph{-fold} cross validation with standard deviation among individual values for each fold.}
\label{tbl:comparison-of-classifiers-signal-noise}
\centering
\begin{tabular}{m{0.6in}m{0.4in}m{0.4in}m{0.6in}m{0.7in}}
\toprule
Frequency bands included & Batch size & \# of hidden units & Best classifier & Validation score \\
\midrule
$\Delta$, H$\alpha$ & 5 & 2 & Multi Layer Perceptron & 0.736 (std: 0.184)\\ \hline
$\Delta$, L$\alpha$, H$\alpha$ & 5 & 4 & Multi Layer Perceptron & 0.736 (std: 0.050)\\ \hline
$\Delta$, $\Theta$, L$\alpha$, H$\alpha$ & 5 & 3 & SVM with linear kernel & 0.755 (std: 0.034)\\ \hline
$\Delta$, $\Theta$, L$\alpha$, H$\alpha$ & 1 & 2 & Gaussian Naïve Bayes & 0.833 (std: 0.068)\\ \hline
$\Delta$, $\Theta$, L$\alpha$, H$\alpha$, L$\beta$ & 5 & 4 & Multi Layer Perceptron & 0.774 (std: 0.098)\\ \hline
\rowcolor{Gray}$\Delta$, $\Theta$, L$\alpha$, H$\alpha$, L$\beta$, H$\beta$ & 10 & 5 & Multi Layer Perceptron & 0.849 (std: 0.058)\\ \hline
$\Delta$, $\Theta$, L$\alpha$, H$\alpha$, L$\beta$, H$\beta$, L$\gamma$ & 5 & 9 & Random Forest & 0.792 (std: 0.075)\\
\bottomrule
\end{tabular}
\end{table}

We began experiments with finding a best validation score for a specific set of auto-encoder's hyper-parameters when only two frequency bands were included in the input data corpus and continued until all bands were involved. We found that in all experiments the classification pipeline based on contractive auto-encoder was superior to that based on the denoising auto-encoder. Therefore, we will provide only the best results for the contractive auto-encoder together with the classifier name and the included frequency bands (see Table~\ref{tbl:comparison-of-classifiers-signal-noise}).

The best signal / noise classification score ($0.849$) obtained for the machine learning classification pipeline, including the contractive auto-encoder, followed by the classifier based on multi-layer perceptron architecture. See Table~\ref{tbl:hyper-parameters-auto-encoders},~\ref{tbl:comparison-of-classifiers-signal-noise} for used auto-encoder hyper-parameters. Optimum hyper-parameters of the most effective classifier (MLP) - momentum: $0.999$, activation: 'relu', solver: 'sgd', alpha: $0.0001$, learning\_rate\_init: $0.001$, hidden\_layer\_sizes: $8$

As a result of the experiments, we found that for the signal / noise classification the SSVEP response elicited by visual stimulation is not important. Only the level of the control signal (Attention or Meditation) during the recording of EEG events is of primary importance. The best result was achieved when almost all frequency bands is included in the analyzed data corpus, which confirms that the elicited by visual simulation SSVEP response is not important for this task.

Thus, we can conclude that the proposed method of signal / noise classification can be used without any visual stimulation. The scope of this method application can be related both to the monitoring of EEG hardware failures and to control the user's level of consciousness in intensive care units.


\subsection{The Classification of Participants}

The second goal of our research is to build a model of machine learning classification pipeline that allows correctly labeling EEG records for specific participants, that is, allowing to classify a participant on the basis of her unique fingerprint in the collected EEG data.

In our experiments we decided to apply visual stimulation with main flicker frequency at $10$Hz and secondary - at $5$Hz, which corresponds to Theta ($\Theta$) and High Alpha (H$\alpha$) bands. In accordance with \cite{Ding-Sperling-Srinivasan:2006}, we assumed that elicited SSVEP response should be detected at these or adjacent frequency bands and that the best validation score will be obtained if only these frequency bands are included in the input data corpus.

A series of experiments was performed on the collected signal-only EEG records with various frequency bands included and the best result ($0.800$) was achieved with data corpus consisting of signal-only records with Delta ($\Delta$) and High Alpha (H$\alpha$) bands included. It almost perfectly fits into the expected range - the Delta ($\Delta$) band is adjacent to the Theta ($\Theta$) band. This congruence is a good indication that SSVEP stimulation is important when collecting EEG records to be classified by participants. See Table~\ref{tbl:comparison-of-classifiers-users} for details about the best results found for various data corpus and hyper-parameters configurations.

\begin{table}[ht]
\renewcommand{\arraystretch}{1.5}
\caption{Comparison of the best validation scores for signal-only classification per particular set of included frequency bands and specific values of hyper parameters of contractive auto-encoder. The score value is given as an average of $3$\emph{-fold} cross validation with standard deviation between individual values for each fold.}
\label{tbl:comparison-of-classifiers-users}
\centering
\begin{tabular}{m{0.6in}m{0.4in}m{0.4in}m{0.6in}m{0.7in}}
\toprule
Frequency bands included & Batch size & \# of hidden units & Best classifier & Validation score \\
\midrule
$\Theta$, H$\alpha$ & 5 & 3 & Multi Layer Perceptron & 0.750 (std: 0.165)\\ \hline
\rowcolor{Gray}$\Delta$, H$\alpha$ & 5 or 1 & 2 & Random Forest & 0.800 (std: 0.041)\\ \hline
$\Delta$, L$\alpha$, H$\alpha$ & 5 & 2 & Multi Layer Perceptron & 0.750 (std: 0.065)\\ \hline
$\Delta$, $\Theta$, L$\alpha$ & 5 & 2 & Multi Layer Perceptron & 0.600 (std: 0.267)\\ \hline
$\Delta$, $\Theta$, L$\alpha$, H$\alpha$ & 5 & 5 & Random Forest & 0.650 (std: 0.138)\\ \hline
$\Delta$, $\Theta$, L$\alpha$, H$\alpha$, L$\beta$ & 5 & 4 & Random Forest & 0.700 (std: 0.135)\\ \hline
$\Delta$, $\Theta$, L$\alpha$, H$\alpha$, L$\beta$, H$\beta$ & 1 & 5 & Random Forest & 0.750 (std: 0.102)\\
\bottomrule
\end{tabular}
\end{table}


\section{Future Works}\label{sec:future-works}


Our current research based on very limited data corpus collected from only three participants. Because of this limitation, we may have missed some important patterns in EEG records, which can further improve the validation scores of studied machine learning classification pipeline methods. In the future, we plan to collect and process EEG records from a much wider audience of participants.


\section{Conclusion}\label{sec:conclusion}


In this paper, we demonstrated how to build machine learning classification pipeline capable of processing and reliably classifying EEG records collected from different participants. We described a visual stimulation scheme suitable to elicit specific SSVEP response from visual cortex and hypothesized how to measure its quality using the collected EEG recordings. In addition, special attention was paid to creating an aesthetically attractive scheme of visual stimulation in the assumption of its use in consumer devices.

We managed to get a fairly good estimate of confidence score with signal / noise classification task (best validation score: $0.849$, see Table~\ref{tbl:comparison-of-classifiers-signal-noise}). This confirms that the proposed method is suitable for automatic processing of EEG records, when it is important to distinguish the signal from noise.

We also demonstrate that, using the proposed method, it is possible with sufficient confidence to classify EEG records for each participant (best validation score: $0.800$, see Table~\ref{tbl:comparison-of-classifiers-users}). We believe that further improvement of the method can provide sufficiently reliable means for use even for user identification based on the collected EEG data.

\newpage


%

\end{document}